%% file: emnlp2023.tex
\pdfoutput=1

\documentclass[11pt]{article}

\usepackage[]{EMNLP2023}

\usepackage{times}
\usepackage{url}
\usepackage{latexsym}
\usepackage{amsfonts}
\usepackage{amsmath}
\usepackage{graphicx}
\usepackage{multirow}
\usepackage{hhline}
\usepackage{floatrow}
\usepackage{array}
\input{bangla_commands}
\usepackage{hyperref}

\usepackage[T1]{fontenc}

\usepackage[utf8]{inputenc}

\usepackage{microtype}

\usepackage{inconsolata}

\title{Visual Question Generation in Bengali}

\author{Mahmud Hasan, Labiba Islam, 
 Jannatul Ferdous Ruma \\
 {\bf Tasmiah Tahsin Mayeesha} and {\bf Rashedur M. Rahman}\\
 North South University\\
{\tt \{mahmud.hasan03,labiba.islam,jannatul.ruma,} \\ \tt{tasmiah.tahsin,rashedur.rahman\}@northsouth.edu} \\}

\date{}

\begin{document}
\maketitle
\begin{abstract}
The task of Visual Question Generation (VQG) is to generate human-like questions relevant to the given image. As VQG is an emerging research field, existing works tend to focus only on resource-rich language such as English due to the availability of datasets. In this paper, we propose the first Bengali Visual Question Generation task and develop a novel transformer-based encoder-decoder architecture that generates questions in Bengali when given an image. We propose multiple variants of models - (i) image-only: baseline model of generating questions from images without additional information, (ii) image-category and image-answer-category: \textit{guided} VQG where we condition the model to generate questions based on the answer and the category of expected question. These models are trained and evaluated on the translated VQAv2.0 dataset. Our quantitative and qualitative results establish the first state of the art models for VQG task in Bengali and demonstrate that our models are capable of generating grammatically correct and relevant questions. Our quantitative results show that our image-cat model achieves a BLUE-1 score of 33.12 and BLEU-3 score of 7.56 which is the highest of the other two variants. We also perform a human evaluation to assess the quality of the generation tasks. Human evaluation suggests that image-cat model is capable of generating goal-driven and attribute-specific questions and also stays relevant to the corresponding image.

\end{abstract}

\section{Introduction}

Visual Question Generation (VQG) is an emerging research field in both Computer Vision and Natural Language Processing. The task of VQG simply uses an image and other side information (e.g. answers or answer categories) as input and generates meaningful questions related to the image. Tasks like cross-modal Visual Question Answering (VQA)
\begin{figure}[ht]
      \includegraphics[width=\linewidth]{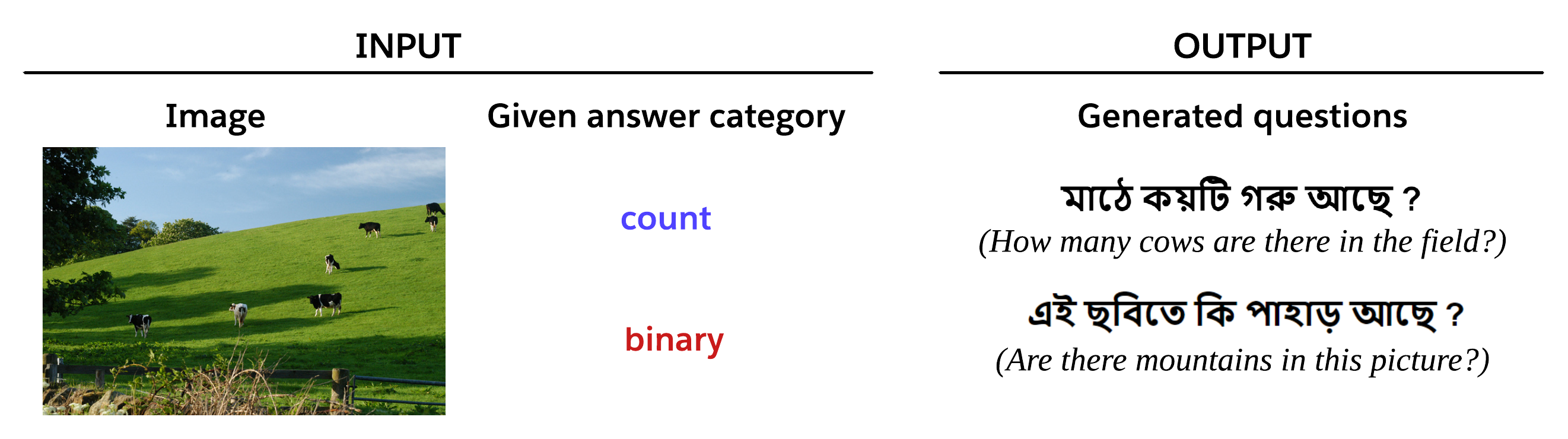}
    \caption{Examples of Bengali VQG Predictions with category of answers as additional information.}
    \label{fig:introduction}
\end{figure} 
\cite{ref1, ref2, ref3, ref4, ref5}, Video Captioning (VC) \cite{ref13}, Image Captioning (IC) \cite{ref6, ref7, ref8}, and Multimodal Machine Translation 
\cite{ref9, ref10, ref11, ref12} are the recent advances in the AI community. While the majority of visuo-lingual tasks tend to focus on VQA, a few recent approaches have been proposed, focusing on the under-researched multi-modal task of VQG. VQG is a more creative and particularly challenging problem than VQA, because the generated questions need to be relevant, semantically coherent and comprehensible to the diverse contents of the given image. 


Existing studies on Visual Question Generation (VQG) have been primarily focused on languages that have ample resources, such as English. While some VQA research have been conducted in low-resource languages like Hindi \cite{ref36}, Bengali \cite{ref38}, Japanese \cite{ref35}, and Chinese \cite{ref37}, limitations have been identified specifically in the context of Bengali language. While Bengali language has some recent work on reading comprehension based question answering \cite{ref46,ref47} and visual question answering \cite{ref38, ref48}, there has been no research conducted for VQG task specifically in Bengali language. 

To obtain meaningful questions, some VQG methods have either augmented the input including additional information such as answer categories, objects in image and expected answers \cite{ref14,ref15,ref16}. \citet{ref14} used ground truth answer with the image as an input, underscoring it to be an effective approach to produce non-generic questions. \citet{ref15} stated that knowing the answers beforehand simply defeats the purpose of generating realistic questions since the main purpose of generating a question is to attain an answer. Instead, they introduced a variational auto-encoder model, which uses the concept of latent space, providing answer categories to generate relevant questions. \citet{ref16}, recently, proposed a guiding approach with three variant families that conditions the generative process to focus on specific chosen properties of the input image for generating questions. Inspired by previous work, we also use additional information such as answer and answer categories in our experiments. To summarize, the main contributions of our paper are the following: 

\begin{itemize}
    \item In our study, we introduce the first visual question generation system that leverages the power of Transformer-based encoder-decoder architecture for the low resource Bengali language. 
    \item We conduct experiments of multiple variants considering only the image and also additional information as input such as answers and answer categories.
    \item We evaluate our novel VQG system with well-established text generation evaluation metrics and report our results as the state of the art in Visual Question Generation in Bengali.
    \item We perform a human evaluation on our generations to assess the quality and the relevance of the questions. 
\end{itemize}

\section{Related Works}


The advent of visual understanding has been made possible due to continuous research in question answering and the availability of large-scale Visual Question Answering (VQA) datasets \cite{ref1, ref27, ref28}. In the past few years, many methods have been proposed to increase the model's performance for a VQG task. Earlier studies \cite{ref8, ref20, ref25, ref32, ref33, ref34} have explored the task of visual question generation through Recurrent Neural Network (RNN), Generative Adversarial Network (GAN), and Variational Auto-Encoder (VAE) which either followed algorithmic rule-based or learning-based approach.

In the visual-language domain, the first VQG paper proposed by \citet{ref25} introduced question-response generation that takes meaningful conversational dialogues as input to generate relevant questions. \citet{ref17} used an LSTM-based encoder-decoder model that automates the generation of meaningful questions with question types to be highly diverse. Motivated by the discriminator setting in GAN, \citet{ref19} formulated a visual natural question generation task that learns two non-generic textual characteristics from the perspective of content and linguistics producing non-deterministic and diverse outputs. Whereas, \citet{ref20} followed the VAE paradigm along with LSTM networks instead of GAN to generate large set of diverse questions given an image-only input. During inference, their obtained results nevertheless required the use of ground truth answers. To defeat this non-viable scenario, \citet{ref15} proposed a VAE model that uses the concept of latent variable and requires information from the target, i.e. answer categories, as input with the image during inference. Similarly, \citet{ref16} follows the concept of latent variable, however, their proposed model architecture explores VQG from the perspective of guiding, which involves two variant families, explicit and two types of implicit guiding approach. Our work is closely related to their explicit guiding method excluding the use of latent space. Recently, \citet{ref31} proposed a BERT-gen model which is capable of generating texts either in mono or multi-modal representation from out of the box pre-trained encoders.

\begin{figure*}[ht]
    \begin{center}
      \includegraphics[width=\textwidth]{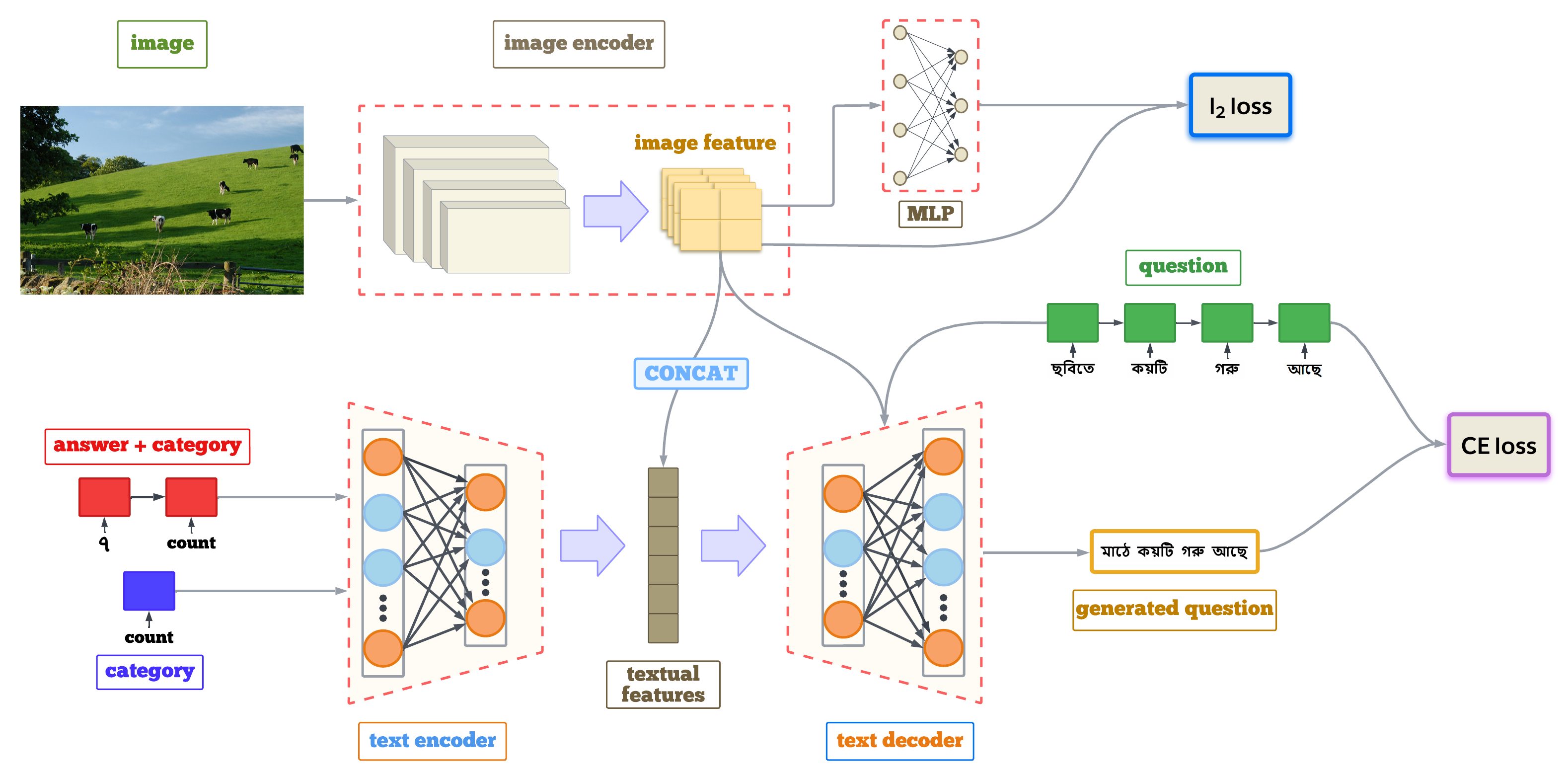}
    \end{center}
    \caption{Architecture of the Bengali VQG Model: Given an image, we first extract image features using an image encoder (CNN). Concatenated form answer and category (\texttt{image-ans-cat}) or only category (\texttt{image-cat}) are given as input to the text encoder to obtain textual features which are then concatenated with the obtained image features. Then, this concatenated form of vision and textual modalities combined with target questions are given as inputs to the decoder question generation in Bengali. Finally, we optimize the CE and MSE loss.}
    \label{fig:training}
  \end{figure*}

\section{Methodology}

In this section, we introduce our transformer based Bengali Visual Question Generation models which can generate meaningful non-generic questions when shown an image along with additional textual information. Our VQG problem is designed as follows: Given an image $\tilde{i}$ $\in$ \emph{I}, where \emph{I} denotes a set of images, decode a question \emph{q}. For each image $\tilde{i}$, we also have access to textual utterances, such as ground truth answer and answer categories. Note, we will use terms "answer category" and "category" interchangeably throughout the paper. In our work, we used answer categories from \cite{ref15} that take 1 out of 16 categorical variables to indicate the type of questions asked. For example, if our model wants to understand answers of "{\bng roNNG} (\textit{color})" category, then it should generate a question ``{\bng ETa ik roeNGr bas ?} (\textit{What color is the bus?})".

Our baseline is an $image-only$ model with no additional textual information like answer or category. We present further two variants both of which shares the same architecture but takes different inputs in training. We feed two different textual  information to our model during training. The first model is $image-ans-cat$ that feeds the concatenated ground truth answer and category to the encoder and is concatenated with the image features. The second model is $image-cat$ that takes only the relevant answer category as input to the encoder. In both of the versions, the input image is reconstructed to maximize information between the image and encoded outputs.

\textbf{Vocabulary:} We construct vocabulary considering all the textual utterances: questions, answers and answer categories. Our vocabulary has a total of 7081 entries including the special tokens. We use word level tokenization. We set a default length of 20 token to each of the questions and 5 to each of the answers. In table \ref{tab:table 1}, we see as maximum length of question in our training dataset is 22 and validation is 21 tokens long, we choose 20 to be the default length. Questions longer than default are truncated and the shorter ones are padded with special \texttt{<pad>} token.

\textbf{Image Encoder:} Given an image $\tilde{i}$, we can extract image features, \emph{f} $\in$ $\mathbb{R}^{\emph{B} \times {300}}$ where B is batch size. Our image encoder is a ResNet-18 pretrained CNN model, which is a convolutional neural network with 72-layer architecture consisting 18 deep layers \cite{ref29}. Once obtaining these features, they are passed to a fully connected layer followed by a batch normalization layer. Specifically, given \emph{f} from image $\tilde{i}$: \emph{i} = ${BatchNorm}(\emph{f})$ $\in$ $\mathbb{R}^{\emph{B} \times {300}}$. 
 
\textbf{Encoder:} We build a Transformer encoder \cite{ref30} and use Bengali pretrained GloVe (Global Vectors for Word Representation) word vectors \cite{ref41} as the embedding layer of the text encoder. Next, we provide answer or answer categories  and image features $f$ as input to the text encoder. Note that, \textit{image-cat} variant only takes answer category $c$ as its input during training and \textit{image-ans-cat} takes concatenated version of answer and category, $[a;c]$ (; operator represents concatenation) as seen in figure \ref{fig:training}. For \textit{image-ans-cat} variant, a concatenated version of answer and category $[a;c]$ is passed through the embedding layer and projected out as context, $C_{img+ans+cat}$ = $ embedding([a;c]) \in \mathbb{R}^{\emph{B} \times \emph{T} \times {300}}$ where, B is batch size and T is the length of the $[a;c]$. For the \textit{image-cat} variant, we only pass the category, $c$ and similarly generate a context $C_{img+cat}$ = $ embedding(c) \in \mathbb{R}^{\emph{B} \times \emph{T} \times {300}}$, where T is the length of $c$.

Additionally, we generate padding masks on answer and category $[a;c]_m = generate-mask([a;c]) \in \mathbb{R}^{\emph{B} \times 1 \times {T}}$ to avoid \texttt{<pad>} tokens being processed by the encoder as well as the decoder. Same operation is performed on category input $c$ and a masked category is generated $c_m$. The \textit{image-cat} model takes context, $C$ and masked category $c_m$ as input to the encoder to encode textual feature representation: $S = encoder(C_{img+cat}, c_m) \in \mathbb{R}^{\emph{B} \times \emph{T} \times {300}}$. We follow the same procedure for the \textit{image-ans-cat} model, where now encoder takes the context, $C_{img+ans+cat}$ and masked concatenated answer and category, $[a;c]_m$. 

These textual feature representation $S$ from the encoder are then concatenated to the input image features $i \in \mathbb{R}^{\emph{B} \times {300}} $, thus, representing our final encoder outputs as the concatenation (; operator) of textual and vision modality: $ X = [S;i] \in \mathbb{R}^{\emph{B} \times \emph{T} \times {300}}$ where B is the batch size and T is length of S. 

\textbf{Decoder:} Our decoder is a Transformer decoder that also uses GloVe embeddings. Following sequence-to-sequence causal decoding practices, our decoder receives encoder outputs from text encoder and ground truth questions during training. We, initially, extract \texttt{<start>} (Start of Sequence) token from encoder outputs which is then taken to the GPU. Each target question is concatenated with a \texttt{<start>} token, forming a tensor. 

In our decoder we follow similar steps as we did in our text encoder. We take ground truth questions $q$ and generate target context: $C_q \in \mathbb{R}^{\emph{B} \times \emph{T} \times {300}}$  and question masks: $q_m \in \mathbb{R}^{\emph{B} \times 1 \times {300}}$.  
Before, we pass the target context, $C_q$ to the decoder, we concatenate it with the same image features $i$ that were passed as input to the encoder previously. The final target context can be denoted by $Q=[C_q;i] \in \mathbb{R}^{\emph{B} \times \emph{T} \times {300}}$. Finally, the decoder takes the encoder outputs $X$ from the text encoder, the concatenated target context $Q$ and the source mask ($[a;c]_m$ or $c_m$ ) depending on the model variant and target question $q_m$ in the form of a tuple. Our decoder is represented as following: $\hat{q}$ = $Decoder(X, Q)$ where the decoder outputs a generated question $\hat{q}$.

\section{Experiments}

\subsection{Datasets}

To collect all relevant information for the VQG task in Bengali, we use the VQA v2.0 \cite{ref1} dataset consisting of 443.8K questions
from 82.8K images in the training dataset, and 214.4K questions from 40.5K images for validation dataset. From the annotations of previous work \cite{ref15}, 16 categories were derived from the top 500 answers. The top 500 answers cover around 82\% of the total VQA v2.0 dataset \cite{ref1}. The annotated categories include objects (e.g. ``{\bng ibrhal} cat", ``{\bng ful} flower", attributes (e.g. ``{\bng Than/Da} cold", ``{\bng puraton} old )", color (``{\bng lal} red", ``{\bng badamii} brown"), etc. 

\begin{table}[ht]
    \centering
    \begin{tabular}{c|c|c}
    \hline
      & Train & Val \\ \hline
      Number of Questions &  184100 & 124795 \\ \hline
      Number of Images & 40800 & 28336 \\ \hline
      Max Length of Question  & 22 & 21 \\ 
      (by words) & & \\ \hline
      Min Length of Question  & 1 & 1 \\ 
      (by words) & & \\ \hline
      Avg Length of Question  & 4 & 4 \\ 
      (by words) & & \\ \hline
    \end{tabular}
    \caption{Analysis of the dataset.}
    \label{tab:table 1}
\end{table}

\begin{figure}[ht]
      \includegraphics[width=\linewidth]{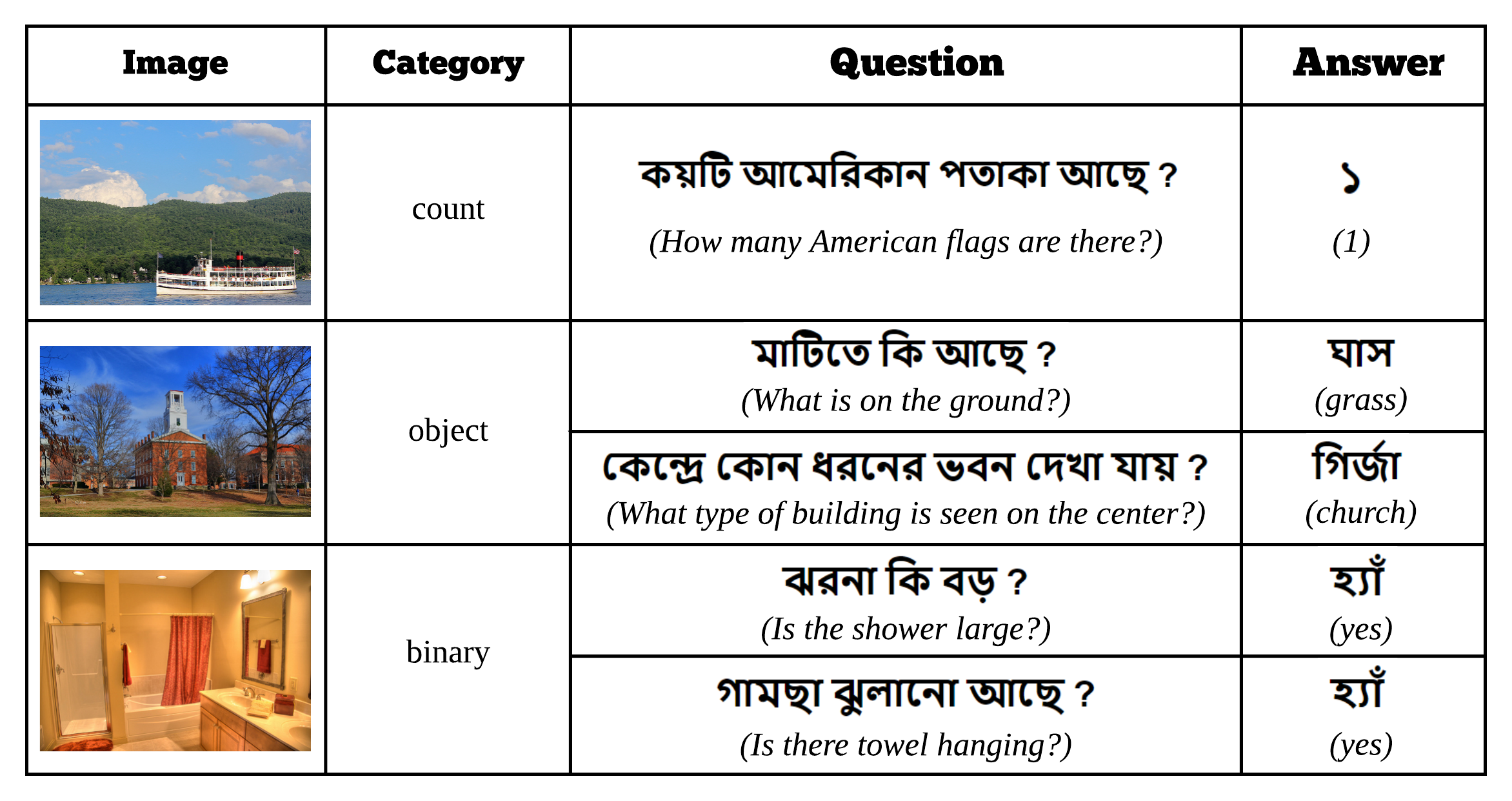}
    \caption{Samples from our dataset}
    \label{fig:dataset}
  \end{figure}

Previously in Bengali machine translation research \cite{hasan-etal-2020-low} , Google translate was found to be competitive with machine translation models trained in Bengali corpora. In another work on Bengali question answering \cite{ref46}, synthetic dataset translated by Google translate was again used for creating Bengali question answering models. Due to Bengali being a low resource language, there has been no available VQG dataset. So we translated the VQA v2.0 \cite{ref1} with Google translate following previous works. We maintained the same partitioning as the original dataset. Due to computational constraints we translated a smaller subset of the training and the validation set. We translate the initial 220K questions and answers for training and 150K questions and answers for validation set in Bengali using GoogleTrans library. In table \ref{tab:table 1}, we see out of 220K training and 150K validation questions, 184K training and 124K validation questions were used. It is because these sets of questions map to top 500 answers in the dataset and we could not use questions and answers that had no mappings to the 16 categories.  In figure \ref{fig:dataset}, we can see the samples of our dataset. The 16 categories in our dataset are  following in English - \textit{``activity'', ``animal'', ``attribute'', ``binary'', ``color'', ``count'', ``food'', ``location'', ``material'', ``object'', ``other'',``predicate'', ``shape', ``spatial'', ``stuff'', ``time''}. 

\begin{figure*}[ht]
    \begin{center}
      \includegraphics[width=\textwidth]{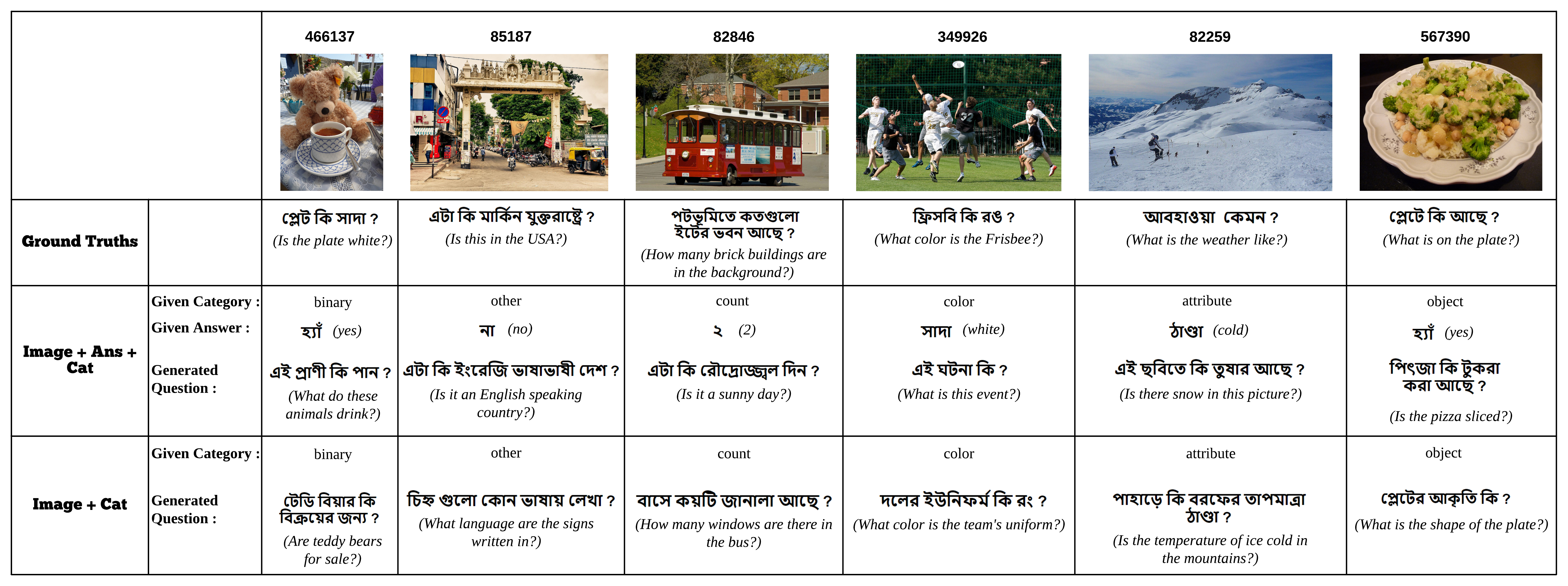}
    \end{center}
    \caption{Qualitative Examples. Ground truths are target questions for both models.}
    \label{fig:qualitative}
  \end{figure*}

\subsection{Training and Optimization}

Our transformer based encoder-decoder architecture is a variation of explicit guiding variant established by \cite{ref16} where  object labels, image captions and object detected features were used as guiding information. However, we only use answer categories and answers as additional information in our work. Instead of BERT \cite{ref21} we use Bengali Glove embeddings \cite{ref49, ref41} for encoding text. We use less number of layers, attention heads and our embedding dimensions and hidden state dimensions are also reduced due to computational constraints. Similar to work done by \cite{ref15} we use the concept of answer category as our primary textual information and attempt to generate questions that are conditioned towards a specific category.

In summary, we begin by first passing the image through a Convolutional Neural Network (CNN) to attain a high dimensional encoded representation of image features, $i$. The image features are passed through an MLP (Multi-layer Perceptron) layer to get a vector representation of reconstructed image features, $i_r$. Our architecture takes an image and additional information in the form of a concatenated answer and category $[a;c]$ or answer category $c$ as input. We feed these input to our text encoder which then generates the textual $S$ and concatenates the textual $S$ and vision modality representations $i$. Our decoder takes the concatenated form of target context $Q$, the encoder outputs $X$, and generates the predicted question, $\hat{q}$ as shown in equation \ref{eq:Decoder}. 

During training, we optimize the $L_q$ between the predicted $\hat{q}$ and target question $q$. Additionally, we try to reconstruct the input image from the encoded output, $X$ and minimize the  $l_2$ loss between the reconstructed image features, $i_r$ and the input image features $i$ to maximize mutual information between the input image features and the encoder outputs as mentioned in equation \ref{eq:loss-function}.     
\begin{equation}
    \hat{q} = Decoder([S;i], [C_q;i])
    \label{eq:Decoder}
\end{equation}
\begin{equation}
    \begin{split}
      L_q & = CrossEntropy(\hat{q}, q) \\
     L_i & = || i - i_r ||_2
\end{split}
\label{eq:loss-function}
\end{equation}

\begin{table*}[h]
\centering
\begin{tabular}{cc*{3}ccc}
\hline
 & \multicolumn{3}{c}{BLEU} & CIDEr & METEOR & ROUGE-L \\ 
\hhline{~---} 
\multirow{-2}{*}{Model} 
 & 1 & 2 & 3 & & \\
\hline
& & \multicolumn{2}{c}{\texttt{ablations}} & & & \\
image-only & 34.84  & 8.04 & 3.98  & 10.62 & 17.14  & 36.56   \\
text-only & 28.05  & 7.57 & 3.65 & 18.72 & 19.10  & 29.68   \\
without-image-recon & 11.59  & 4.85 & 2.08  & 26.61 & 12.34  & 31.43   \\
\hline
& & \multicolumn{2}{c}{\texttt{variants}} & & & \\
image-cat & 33.12 & \textbf{13.52} & \textbf{7.56}  & 22.76 & 17.18 & 36.12  \\
image-ans-cat & 32.97 & 11.80 & 3.82  & 18.63 & 18.63 & 36.90 \\
\hline
\hline
\end{tabular}
\caption{\label{tab:table 2}
Evaluation results of model variants and ablations.
}
\end{table*}


\subsection{Inference}
During inference, except the \texttt{image-only} variant, both model variants are provided with only answer category (e.g. `` {\bng roNG} (\textit{color})", ``{\bng oibishs/TY} (\textit{attribute})", ``{\bng gonona} (\textit{count}"), etc.)  alongside an image during inference, because providing answers to the model would violate the realistic scenario \cite{ref15, ref16}. As a result our model is kept under a realistic inference setting by not providing an answer as input during inference.


\subsection{Evaluation Metrics}
In our experiments, we followed well established language modelling evaluation metrics BLEU: \cite{ref42}, CIDEr \cite{ref43}, METEOR \cite{ref44}, and ROUGE-L \cite{ref45}. 

\subsection{Implementation details}
We use a pretrained ResNet18 as our image encoder to encode image features. Both of our transformer based encoder and decoder uses glove embeddings. We set our transformer encoder and decoder with the following setting: number of layers = 4, number of attention heads = 4, embedding dimension = 300, hidden dimension = 300 and filter size = 300. The model trains a total number of 13000 steps, with a learning rate of 0.003 and batch size of 64. We have implemented our model with pytorch. We expect to release our code and translated dataset publicly at \href{https://github.com/mahmudhasankhan/vqg-in-bengali}{https://github.com/mahmudhasankhan/vqg-in-bengali}

\begin{table*}[h]
\centering
\begin{tabular}{cc*{3}ccc}
\hline
 & \multicolumn{3}{c}{Experiment} \\ 
\hhline{~---} 
\multirow{-2}{*}{Model} 
 & 1 & 2  \\
\hline
image-cat & 47.5\% & 40\%    \\
image-ans-cat & 30\% & 37.5\%  \\
\hline
\end{tabular}
\caption{\label{tab:table 3}
Human evaluation result of our model variants.
}
\end{table*}

\subsection{Model Ablations}
We experiment with a series of ablations performed on our model such as \texttt{image-only} does not include text encoder. Inversely, \texttt{text-only} model does not have image encoder. With respect to \texttt{without-image-recon}, we avoid optimizing the reconstruction $l_2$ loss between the reconstructed image features and input image features. As for our model variants, \texttt{image-cat} and \texttt{image-ans-cat}, the entire architecture remains intact.

\section{Results}

\subsection{Quantitative Results}
We test our model variants except with only categorical information because giving answer to a model beforehand would be unrealistic. We tried to figure out which textual input is more significant and leads to better results. Firstly, our model ablations justify our model architecture as such our intact architecture outperforms all the ablations in BLEU-2 and BLEU-3 (see Table \ref{tab:table 2}). Our baseline \texttt{image-only} model achieves a BLEU-3 score of 3.98 which is higher than \texttt{image-ans-cat} variant. Moreover, we find that in some metrics \texttt{image-cat} model outperforms the \texttt{image-ans-cat} model and in some metrics stay ahead marginally. As seen in table \ref{tab:table 2}, \texttt{image-cat} model achieves a BLEU-3 score of 7.56 that is almost 4 points ahead of both \texttt{image-only} and \texttt{image-ans-cat} model. Moreover, we notice that \texttt{image-cat} model also performs marginally better in CIDEr metrics. However, both the variants show similar performance on other evaluation metrics except for METEOR and ROUGE-L metric where \texttt{image-ans-cat} variant performs slightly better. In comparison to \cite{ref16} for experiments in explicit image-category setting for English, our BLEU-1 score is 33.12 while for English we see a score of 40.8 with a 7.68 difference, however, BLEU-2 and BLEU-3 scores have higher differences. However, for METEOR in English, the score is 20.8 while our \texttt{image-cat} model scores 17.18 with a 3.62 difference only and for ROUGE the English score is  43.0 while we score 36.12 with a 6.88 point difference. Similar experiments on guided visual generation have not been performed for other languages or Bengali to our knowledge, so we compare only with English. 
While our scores are lower than English, we train on smaller and translated dataset for computational and data annotation related constraints. Based on the quantitative results we can come to a conclusion that categorical information shows better results overall. In the next section, we see the qualitative results where we shall see that categorical information conditions the \texttt{image-cat} variant to generate category specific questions i.e. goal driven, attribute specific questions rather than generic questions.

\subsection{Qualitative Results}

In figure \ref{fig:qualitative}, we can compare the generated questions from our model variants with the reference ground truth question and answer category more illustratively. Questions generated from the  \texttt{image-cat-ans} model although are grammatically and semantically correct but in some cases are not conditioned towards the given category. For example, in image 82846, although the question is grammatically correct, however, the generated question does not follow the given category which is ``count''. We see similar behavior for images 349926 and 82259 where questions are grammatically correct and relevant to the image but do not follow the category. In contrast, the \texttt{image-cat} model perfectly conditions its questions towards the given category. The questions are not only grammatically and semantically valid but also follow the given categorical information. The questions from the \texttt{image-cat} model generates goal driven, non-generic and category oriented questions. To understand why this variant of VQG performs well although having less side information during training, is likely due to the fact that in validation step both variants only take category side information. Therefore, the \texttt{image-cat} learns better than \texttt{image-ans-cat}.

Additionally, we notice that both variants are able to decode the semantic information from the input image as well. Both variants can rightly identify the objects and features present in the images.
\subsection{Human Evaluation}

We conducted a human evaluation to understand the quality of the generated questions similar to work done in \cite{ref16}. In our experiments, we ask three annotators to evaluate our generated questions with two questions. There was no annotator overlap where two annotators annotated the same question. We evaluate category wise question generation by comparing two of our model variants, \texttt{image-cat} and \texttt{image-ans-cat}. 

In \textit{Experiment 1}, known as the Visual Turing Test, we present annotators with an image, a ground truth question, and a model-generated question. The task of annotators is to discern which question, among the two, they think is produced by the model. \textit{Experiment 2} involves displaying an image to the annotators along with a question generated by the model. Subsequently, the annotators are asked to decide whether the generated question seems relevant to the given image. For each of the experiments we annotate 40 generations for each models, resulting in 80 annotations per experiment. The complete results of our evaluation is listed in table \ref{tab:table 3}.

In \textit{Experiment 1}, the result of our \texttt{image-cat} model outperforms the \texttt{image-ans-cat} variant fooling humans about 47.5\% of the time. In a Visual Turing Test, if a model is capable of generating human-like questions, it is expected that its performance would reach approximately 50\%. Although close to the desired score of 50\%, the \texttt{image-cat} variant represents a promising advancement in surpassing the Visual Turing Test. We evaluate \textit{Experiment 2} on both our model variants where the \texttt{image-ans-cat} model shows a percentage score of 37.5\%, outperforming the \texttt{image-cat} model. It is possible that providing the answer with the image and the category helps in generating more relevant questions. 

\section{Conclusion}
We proposed the first VQG work in Bengali and presented a novel transformer based encoder-decoder architecture that generates questions in Bengali when shown an image and a given answer category. In our work, we presented two variants of our architecture: \texttt{image-cat} and \texttt{image-ans-cat} that differs from what input they receive during training. Both of the variants generate a question based on answer category as guiding information from an image. However, due to having two different input combinations, \texttt{image-cat} performs marginally better in terms of quantitative scores, however, generates goal driven, specific questions conditioned towards the categorical information it receives. In contrast, the \texttt{image-ans-cat} model although generating grammatically valid questions fail to learn about answer categories. Future work could analyze the impact of using more modern CNN architectures and newer pretrained models to generate questions from images.

\section{Acknowledgement}
This work was funded by the Faculty Research Grant [CTRG-22-SEPS-07], North South University, Bashundhara, Dhaka 1229, Bangladesh

\bibliography{anthology,custom}
\bibliographystyle{acl_natbib}




\end{document}

%% file: bangla_commands.tex
\def\bng{\bngx}

\font\bngx=bang10

\def\*#1*#2{o\null{#2}{#1}}

\def\sh#1{\setbox0=\hbox{#1}%
     \kern-.02em\copy0\kern-\wd0
     \kern.04em\copy0\kern-\wd0
     \kern-.02em\raise.0433em\box0 }